\documentclass[10pt,twocolumn,twoside]{IEEEtran}

\ifCLASSOPTIONcompsoc
  \usepackage[nocompress]{cite}
\else
  \usepackage{cite}
\fi

\ifCLASSINFOpdf
  \usepackage[pdftex]{graphicx}
\else
  \usepackage[dvips]{graphicx}
\fi

\usepackage{amsthm}
\usepackage{amsmath}
\usepackage{bm}
\usepackage{multirow}
\usepackage{enumitem}
\usepackage{hyperref}       
\hypersetup{
    colorlinks=true,
    linkcolor=blue,
    urlcolor=blue,
    citecolor=purple,
}
\usepackage{url}            
\usepackage{booktabs}       
\usepackage{amssymb}
\usepackage{makecell}
\usepackage{color}
\usepackage{mathrsfs}
\usepackage{pifont}
\usepackage[table,xcdraw]{xcolor}

\usepackage{wrapfig}

\usepackage{setspace}
\usepackage{color}
\ifCLASSOPTIONcompsoc
  \usepackage[caption=false,font=normalsize,labelfont=sf,textfont=sf]{subfig}
\else
  \usepackage[caption=false,font=footnotesize]{subfig}
\fi

\definecolor{gray}{rgb}{0.35,0.35,0.35}
\definecolor{blue}{rgb}{0,0,1}
\definecolor{red}{rgb}{1,0,0}
\definecolor{orange}{rgb}{0.75, 0.4, 0}
\definecolor{purple}{rgb}{0.5, 0.0, 0.5}
\definecolor{black}{rgb}{0,0,0}

\begin{document}
\title{Manifold-Aware Local Feature Modeling for \\Semi-Supervised Medical Image Segmentation}
\author{Sicheng Shen, Jinming Cao*\thanks{*Corresponding author.}, Yifang Yin, and Roger Zimmermann, \IEEEmembership{Senior Member, IEEE}
\thanks{Sicheng Shen, Jinming Cao and Roger Zimmermann are with National University of Singapore, Singapore. E-mail: sshen@u.nus.edu, jinming.ccao@gmail.com, rogerz@comp.nus.edu.sg.}
\thanks{Yifang Yin is with Institute for Infocomm Research (I$^2$R), A*STAR, Singapore. E-mail: yin\_yifang@i2r.a-star.edu.sg. }}

%



\maketitle
\begin{abstract}
Achieving precise medical image segmentation is vital for effective treatment planning and accurate disease diagnosis. 
Traditional fully-supervised deep learning methods, though highly precise, are heavily reliant on large volumes of labeled data, which are often difficult to obtain due to the expertise required for medical annotations. 
This has led to the rise of semi-supervised learning approaches that utilize both labeled and unlabeled data to mitigate the label scarcity issue. 
In this paper, we introduce the Manifold-Aware Local Feature Modeling Network (MANet), which enhances the U-Net architecture by incorporating manifold supervision signals. 
This approach focuses on improving boundary accuracy, which is crucial for reliable medical diagnosis. 
To further extend the versatility of our method, we propose two variants: MA-Sobel and MA-Canny. 
The MA-Sobel variant employs the Sobel operator, which is effective for both 2D and 3D data, while the MA-Canny variant utilizes the Canny operator, specifically designed for 2D images, to refine boundary detection. 
These variants allow our method to adapt to various medical image modalities and dimensionalities, ensuring broader applicability.
Our extensive experiments on datasets such as ACDC, LA, and Pancreas-NIH demonstrate that MANet consistently surpasses state-of-the-art methods in performance metrics like Dice and Jaccard scores. The proposed method also shows improved generalization across various semi-supervised segmentation networks, highlighting its robustness and effectiveness. Visual analysis of segmentation results confirms that MANet offers clearer and more accurate class boundaries, underscoring the value of manifold information in medical image segmentation.
We open-sourced an implementation of MANet in PyTorch at \href{https://github.com/SichengS/MANet}{\underline{https://github.com/MANet}}.
\end{abstract}


\section{Introduction}
\label{sec:intro}
Precise Medical image segmentation provides insightful information for treatment planning and disease diagnosis~\cite{basak2023pseudo}.
Compared to semantic segmentation tasks in purely indoor~\cite{wu2024point} or outdoor~\cite{chenvision} scenes, medical image segmentation~\cite{basak2023pseudo, wang2024towards, yu2019uncertainty, li2020transformation, 10526382} demands a higher level of detail and shape accuracy since erroneous segmentation of organ details can adversely affect medical staff judgment and potentially lead to medical malpractice. 
Various fully-supervised deep learning methods have been proposed~\cite{chen2017deeplab, milletari2016v, ronneberger2015u}, achieving high precision. 
However, these fully-supervised methods require a large amount of labeled data. 
Unlike datasets for indoor~\cite{armeni20163d,cao2021shapeconv} or outdoor~\cite{cordts2016cityscapes} scene segmentation, medical image data demand high expertise from annotators, such as relevant medical knowledge, which limits the availability of labeled medical images for model training~\cite{chi2024adaptive}.
Consequently, semi-supervised learning~\cite{bai2023bidirectional, wang2023mcf, rahman2023medical, huang2024combinatorial, wu2022exploring} has gained popularity in recent years for medical image segmentation. 
This approach significantly alleviates the pressure of label scarcity by leveraging a small amount of labeled data along with a relatively large portion of unlabeled data.

\begin{figure}[t!]
	\centering
	\centering
	\includegraphics[width=.48\textwidth]{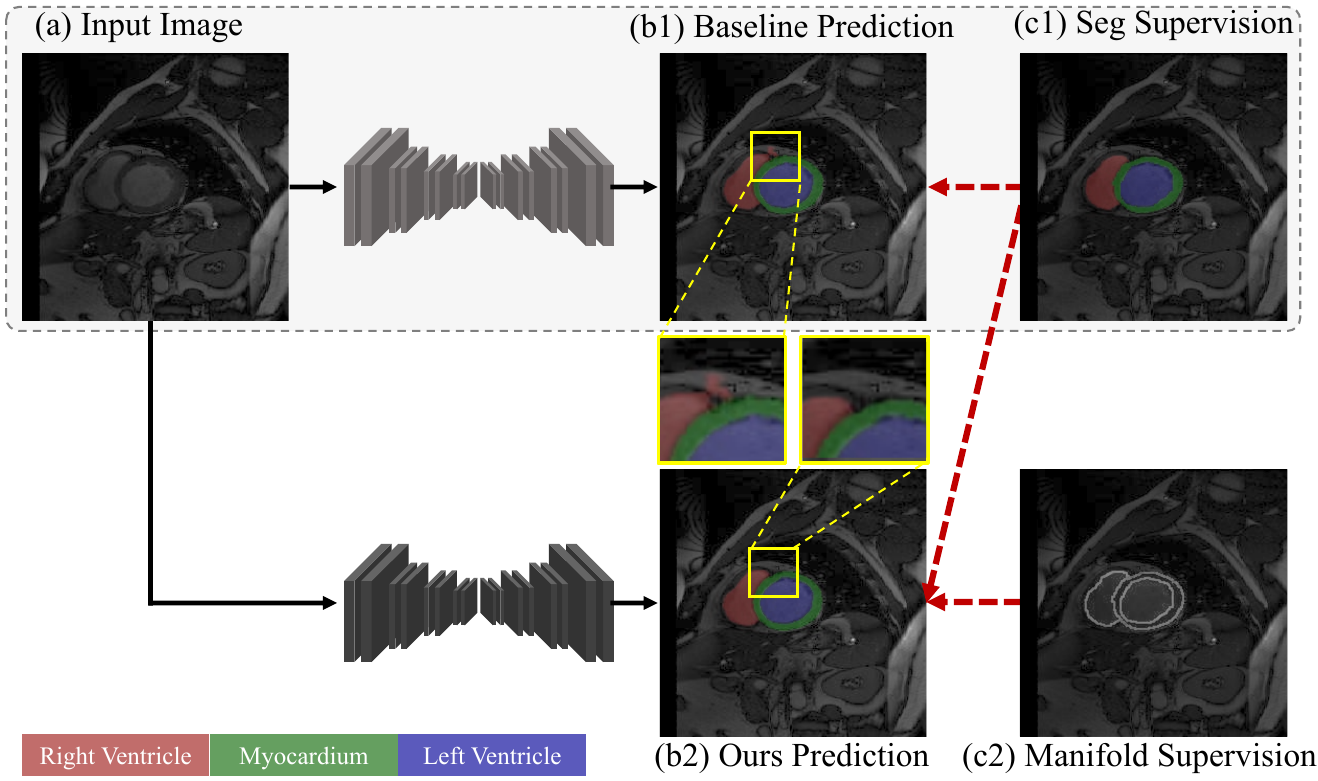}
	\caption{
 Visual demonstration of the importance of manifold supervision in medical imaging. (a) Input image from the ACDC dataset. (b1) Baseline prediction with segmentation supervision. (c1) The supervision signal of the baseline method, including the ground truth for labeled data and pseudo-labels for unlabeled data. (b2) Our prediction with both segmentation and manifold supervision. (c2) The manifold supervision signal. The boundary areas in predictions with manifold supervision are clearer than those in results with segmentation supervision alone.}
	\label{fig:teaser}
\end{figure}

In the field of semi-supervised medical image segmentation, numerous methods are proposed to better utilize the limited amount of labeled data. Pseudo-label generation~\cite{basak2023pseudo,chen2021semi,rizve2021in,wang2022semi} and consistency regularization~\cite{tarvainen2017mean,yu2019uncertainty,ouali2020semi} are two major categories in this area. Pseudo-label generation methods leverage the model’s own predictions on unlabeled data by treating these predictions as pseudo-labels. Once these pseudo-labels are generated, the model is retrained using them, essentially using the model’s inferred knowledge to guide further learning. This approach helps in improving the model’s performance by utilizing all available data, even if it is not originally labeled. Consistency regularization methods, on the other hand, work by applying various augmentations or perturbations to the unlabeled data. The main objective is to ensure that the model produces consistent predictions across different versions of the same data point, despite the changes introduced by these augmentations. By enforcing consistency, the model learns to be robust and stable in its predictions, contributing to its overall accuracy and reliability. 
Both approaches highlight the importance of overall semantic information. 
As shown in Figure~\ref{fig:teaser}, (a) represents the original input medical image, and (b1) shows the predicted result generated by the network trained with segmentation supervision signals (c1). We can observe that the current network model is capable of capturing long-range semantics, with most areas being accurately segmented. However, we notice that there are more pronounced classification errors in the boundary regions between different categories (i.e., the right ventricle cavity and the background) compared to the internal regions of different categories. Although these boundary regions are localized, they are crucial in medical imaging because accurate diagnosis often relies on precise delineation of organ edges and shapes. Nevertheless, current methods typically do not focus on these critical localized areas.

Motivated by this observation, we propose integrating boundary information into the network stream. 
Specifically, for medical image datasets, variations in acquisition methods—such as MRI~\cite{bernard2018deep} and CT~\cite{roth2015deeporgan}—often result in different data dimensionalities, typically 2D and 3D. 
The method proposed in this paper is designed to be broadly applicable across these variations. 
For 3D data, we extend the concept of image boundaries to object surfaces, thereby generalizing boundary information to manifold information.
We refer to our method as the Manifold-Aware Local Feature Modeling Network (MANet), which leverages manifold information as a core component of its design. 
To support various ways of generating manifold information, we provide two effective approaches within the network. 
The first is the Sobel operator~\cite{sobel19683x3}, which is generally applicable to both 2D and 3D data, while the second is the Canny operator~\cite{kovalevsky2022edge}, which is specifically designed for 2D images. Based on these two configurations, we derive two variants of our method: MA-Sobel and MA-Canny, tailored for different data setups.
As shown in Figure~\ref{fig:teaser}, compared to the original single segmentation supervision signal, we introduced a manifold supervision signal (c2) to train the network then produce new prediction results (b2). It is evident that the segmentation results in (b2) provide clearer class boundaries than those in (b1), thereby offering more valuable assistance for subsequent diagnosis.
More specifically, many existing medical image segmentation approaches~\cite{bai2023bidirectional,wang2023mcf,gao2023correlation, huang2024combinatorial, wu2022exploring, 9558816} primarily utilize the U-Net architecture~\cite{ronneberger2015u}, which features an encoder-decoder structure.
The encoder extracts features from the input image, while the decoder upsamples these features to the original image size to generate segmentation predictions.
We propose enhancing the U-Net architecture by adding a Manifold branch. This branch shares the same input features as the original segmentation task but uses manifold information as a supervisory signal. 
By employing a semi-supervised training approach, we leverage both labeled and unlabeled data. 
For the unlabeled data, the supervisory signal for the newly added branch is generated using pseudo-labels by Manifold-Generator. This method allows us to incorporate more supervisory signals without the need for additional labeled annotations.
Moreover, the manifold branch is used only during the training phase. 
During testing, we can still use the updated encoder and the segmentation branch for inference. This approach yields better performance without increasing inference time compared to baseline methods.

To evaluate the effectiveness of our proposed method, we conduct extensive experiments on multiple well-established datasets, including the 2D datasets ACDC~\cite{bernard2018deep}, as well as the 3D dataset LA~\cite{xiong2021global} and Pancreas-NIH~\cite{roth2015deeporgan}. 
Our experimental result shows that MANet outperforms existing state-of-the-art methods~\cite{yu2019uncertainty, li2020shape, luo2021semi, wu2022exploring} in terms of Dice score and Jaccard sore across all settings. To validate its generalization capabilities, we applied our method to various widely adopted semi-supervised segmentation networks~\cite{bai2023bidirectional, wang2023mcf, gao2023correlation}. An improved performance is observed when compared to baselines.
Moreover, through the visualization of the segmentation results, we observed that the segmentation results using MANet exhibit higher accuracy at class boundaries, highlighting the effectiveness of manifold information in medical image segmentation tasks. Code and related parameter settings can be found in the supplementary material to enable others to reproduce this work.

Our main contributions can be summarized as follows: 

1. We proposed MANet, a novel module that utilizes manifold information to provide additional supervision signals to the model and enhance the modeling of local features. This structure allows the network to better focus on manifold region, which is crucial in medical imaging. 

2. We introduced two variants of MANet: MA-Sobel and MA-Canny. MA-Sobel applies the Sobel operator for both 2D and 3D data, providing a versatile solution, while MA-Canny, designed for 2D images using the Canny operator, achieves superior boundary accuracy on 2D datasets.

3. The manifold branch in MANet can be seamlessly integrated into a semi-supervised segmentation baseline network with a U-Net architecture, thereby enhancing the performance of the baseline method. 

4. The additional branch in MANet can be ablated during the inference phase, ensuring that our method does not introduce any additional computational cost compared to the baseline method during testing. This is an important consideration in practical applications.
\section{Related Work}
\label{sec:rw}

\subsection{Semi-supervised segmentation}
\begin{figure*}[t!]
	\centering
	\centering
	\includegraphics[width=0.95\textwidth]{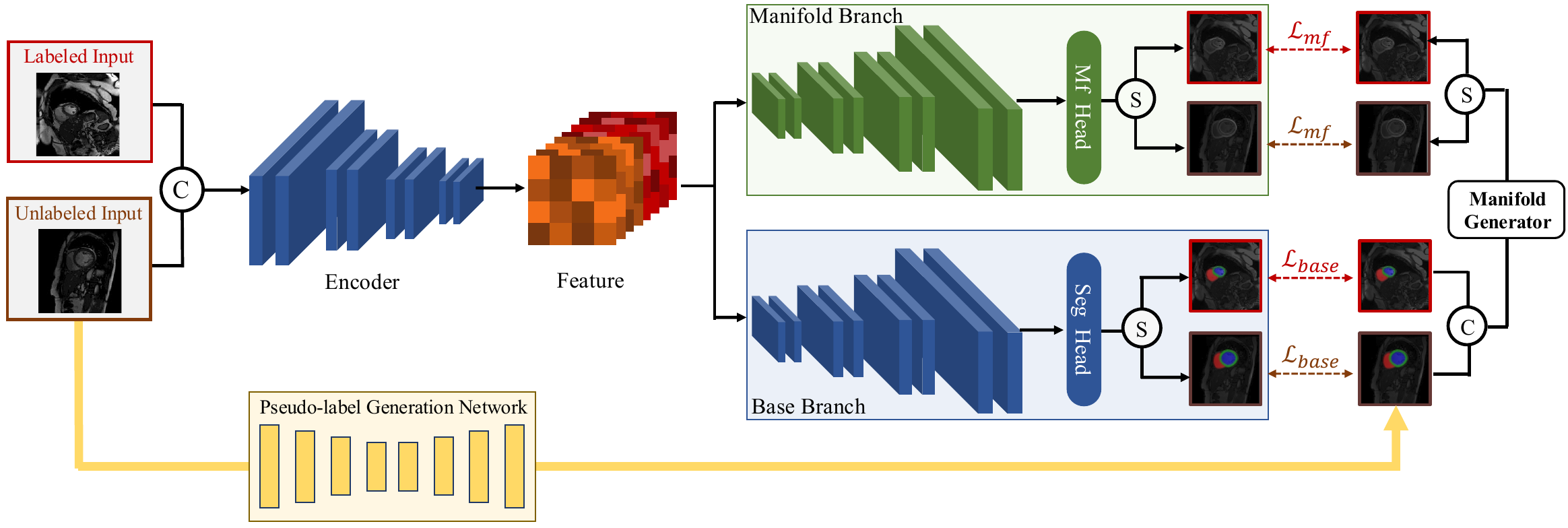}
	\caption{The overall network architecture. In the figure, labeled and unlabeled data are first concatenated batch-wise ($\textcircled{c}$) and then fed simultaneously into the encoder to extract feature representations. These features are passed into the decoder, which consists of two branches: the base branch and the manifold branch. The final layer of each branch is a distinct output head, with the base branch utilizing a segmentation head and the manifold branch employing a manifold head. These heads generate predictions, which are then split batch-wise ($\textcircled{s}$) into labeled and unlabeled portions, supervised by their respective segmentation and manifold signals. Specifically, for unlabeled data, the segmentation supervision is derived from the pseudo-labels generated by the pseudo-label generation network. Meanwhile, the manifold supervision is generated by the manifold generator, which uses the corresponding segmentation signals.  During the testing phase, the manifold branch can be removed, allowing the network to perform inference using only the base branch (highlighted by the blue modules), thereby incurring no additional computational cost compared to the baseline method.}
 
	\label{fig:Architecture}
\end{figure*}

Manual pixel-level annotations for semantic segmentation are very time-consuming and costly. 
Therefore, semi-supervised segmentation~\cite{lee2013pseudo,rahman2023medical,tragakis2023fully,fan2022ucc,Zhou_2023_BMVC, 9779756} has gained popularity recently which aims to utilize supervision on a few labeled data to extract useful representations from a large amount of unlabeled data.

Existing semi-supervised segmentation methods can be broadly categorized into two main approaches: pseudo-labeling methods and consistency regularization methods. The pseudo-labeling methods~\cite{basak2023pseudo,chen2021semi,rizve2021in,wang2022semi} primarily generate pseudo-labels using techniques, such as MixMatch~\cite{berthelot2019mixmatch} and temporal ensembling~\cite{laine2016temporal}, to address the issue of label scarcity. On the other hand, consistency regularization methods~\cite{tarvainen2017mean,yu2019uncertainty,ouali2020semi} enforce the consistency of predictions or intermediate features under various perturbations using techniques like entropy minimization~\cite{grandvalet2004semi} and Virtual Adversarial Training~\cite{miyato2018virtual}.

\subsection{Medical Image Semi-Supervised Segmentation}
In comparison to other types of data segmentation, such as indoor or outdoor scenes, medical images demand higher expertise from annotators, resulting in increased annotation costs. Therefore, the application of semi-supervised methods in medical image segmentation has emerged as a critical concern.

Many previous works explore methods to generate reliable pseudo-labels. Bai~\textit{et al.}~\cite{bai2023bidirectional} combines copy-paste with the Mean Teacher framework. They randomly paste both labeled and unlabeled patches onto each other to generate new data with some supervision. The outputs of the teacher network are seen as pseudo-labels for unlabeled data.~\cite{wang2023mcf} argues the student network limits the traditional teacher network's performance due to the high structural similarity and parameter update method of EMA. Hence, they simultaneously trained two subnets and chose the better-performing subnet to generate pseudo-labels for the other subnet.~\cite{basak2023pseudo} use pseudo-labels to provide additional guidance for contrastive learning to minimize the effect of domain shift on overall performance. Wang~\textit{et al.}~\cite{wang2024towards} conducted knowledge distillation by using a common encoder and two separate decoders to extract pseudo-labels for an additional decoder for prediction only. 

Consistency regularization is another major approach in semi-supervised medical image segmentation. Huang~\textit{et al.}~\cite{huang2024exploring} proposed a general framework that further extracts latent information from the segmentation map. The latent information is supervised by inherent consistency, which aligns latent representation with the ground truth. Yu~\textit{et al.}~\cite{yu2019uncertainty} incorporates uncertainty estimates to handle noisy pseudo-labels from the teacher network and forces the model to focus on confident prediction. \cite{xie2021intra} proposed a dual consistency module that first creates an attention map for common semantics between different images and imposes consistency constraints on attention maps from image pairs to filter out low-confidence attention regions.

The method proposed in this paper belongs to the first category, focusing on how to effectively utilize pseudo-labels. Specifically, it models the manifold space of pseudo-labels, seamlessly integrating with existing approaches that leverage pseudo-labels. This integration aims to assist the main network in achieving enhanced performance.

\subsection{Manifold Applications in Segmentation}

Our proposed approach leverages manifold spaces, which are lower-dimensional representations encoding the intrinsic data structure of the original input, known as the latent space. For 2D data formats, boundaries simplify the corresponding manifold, whereas for 3D data, surfaces serve this purpose. Existing segmentation methods typically utilize manifolds in singular forms, focusing on boundaries in 2D images, as demonstrated by Marmanis et al. \cite{marmanis2018classification}, who integrate class-boundary networks with segmentation networks via simple addition. Similarly, Chen et al. \cite{chen2016semantic} employ domain transform, an edge-preserving filter, to combine edge maps and segmentation maps.

In the domain of medical images, Liu et al. \cite{liu2022mea} extract edge information from early features and incorporate attention mechanisms for guidance. In contrast to these methods, our approach offers universal applicability across both 2D and 3D data, enhancing baseline method performance using respective manifold spaces, while maintaining crucial inference efficiency in practical applications.
\section{Method}
\label{sec:method}

\subsection{Overall Architecture and Notations} 


The network architecture proposed in this paper, as shown in Figure~\ref{fig:Architecture}, consists of four main components: an encoder, a dual-branch decoder, a pseudo-label generation network, and supervision signals. As depicted, the baseline method typically includes the encoder and the base branch of the decoder (the blue portion), which generates segmentation results guided by segmentation maps as supervision signals. 
In contrast, we enhance the decoder by introducing an additional manifold branch, which supplements the model through the integration of manifold supervision. At the core of this architecture is its ability to simultaneously handle both labeled and unlabeled data inputs. These inputs are processed through the encoder to extract feature representations, which are subsequently fed into both the base and manifold branches.
The base branch functions consistently with the baseline method, aiming to receive supervision signals from ground truth labels (for labeled data) and pseudo-labels generated by the pseudo-label generation network (for unlabeled data). However, since the task we address is semi-supervised learning, the manifold branch does not require additional supervision. Instead, it leverages manifold supervision obtained from the segmentation maps via the manifold generator.
The total loss function of the network aggregates the losses from both branches, balancing them with weights to optimize overall segmentation performance.

To ensure clearance and consistency throughout our presentation, we first define the following notations.
The proposed network architecture is designed to handle both 2D and 3D medical images. For 2D medical images, we define the input image as $\textbf{I} \in R^{W \times H \times C}$, where $W$, $H$, and $C$ represent the width, height, and the number of channels of the image $\textbf{I}$, respectively. Typically, for medical images, $C=1$, as grayscale images are commonly used. The goal of medical image segmentation is to predict the label of each pixel, denoted as $\textbf{Y} \in \{0,1,\dots,K-1\}^{W \times H}$, where $K$ is the total number of segmentation classes.
For 3D volumetric medical data, the input is defined as $\textbf{V} \in R^{W \times H \times L}$, where $L$ is the third dimension of the volume, corresponding to the depth or number of slices. The objective is to predict the label for each voxel, represented as $\textbf{Y} \in \{0,1,\dots,K-1\}^{W \times H \times L}$.

In the baseline method, the training dataset $D$ consists of $N$ labeled data ($D_{L}$) and $M$ unlabeled data ($D_{U}$), where typically $N \ll M$, so that $D = D_{L} \cup D_{U}$. 
Accordingly, we denote the segmentation predictions for labeled and unlabeled inputs as $\hat{Y}^l$ and $\hat{Y}^p$, respectively. The labeled predictions, $\hat{Y}^l$, are directly supervised by the ground truth labels, $Y^{l}$, ensuring accurate segmentation outcomes for the labeled data. In contrast, the unlabeled predictions, $\hat{Y}^p$, are refined through the pseudo-label generation network, which provides supervision via pseudo-labels, $Y^{p}_{pseudo}$. This approach allows the network to leverage both labeled and unlabeled data effectively, enhancing its generalization capability.

In our approach, we introduce a manifold supervision signal to guide the manifold branch in predicting the manifold structure, denoted as $\textbf{M}$. For 2D images, $\textbf{M} \in \{0,1\}^{W \times H}$, and for 3D data, $\textbf{M} \in \{0,1\}^{W \times H \times L}$. This signal shares the same spatial dimensions as the segmentation map $\textbf{Y}$, but consists of only two channels, indicating whether a given location belongs to the manifold or not. The manifold supervision signal is generated using a segmentation map-based manifold generator $\mathbb{M}$.
The manifold supervision applies to both labeled and unlabeled data, enhancing the learning process. Specifically, for labeled data, the manifold supervision is represented as ${M}^{l} = \mathbb{M}(Y^{l})$, while for unlabeled data, it is expressed as ${M}^{U} = \mathbb{M}(Y^{p}_{pseudo})$. Correspondingly, the predicted manifold results for labeled and unlabeled data are denoted as $\hat{M}^{l}$ and $\hat{M}^{p}$, respectively.


\begin{table*}[]
\caption{\label{tab:ACDC}
	Performance comparison with state-of-the-art methods on ACDC dataset.($\uparrow$) and ($\downarrow$) indicate better performance with larger and smaller numbers, respectively. \textbf{Bold} text indicates the best performance. An asterisk (*) indicates the performance we reproduced on the respective datasets using the source code provided by the various methods. Note that there may be slight discrepancies from the results reported in the original publications.
 }
 \centering
\scalebox{0.8}{
\begin{tabular}{c|c|cc|cccc}
\hline
                             &                           & \multicolumn{2}{c|}{Scans Used}                          & \multicolumn{4}{c}{Metrics}                                                                                                                   \\ \cline{3-8} 
\multirow{-2}{*}{Method}     & \multirow{-2}{*}{Venue}   & Labeled                    & Unlabeled                   & Dice(\%)↑                              & Jaccard(\%)↑                           & 95HD↓                        & ASD↓                         \\ \hline
U-Net~\cite{ronneberger2015u}                        & -                         & 3 (5\%)                   & 0                           & 47.83                                  & 37.01                                  & 31.16                         & 12.62                         \\ 
U-Net~\cite{ronneberger2015u}                        & -                         & 7 (10\%)                   & 0                           & 79.41                                  & 68.11                                  & 9.35                         & 2.70                         \\
U-Net~\cite{ronneberger2015u}                        & -                         & 70 (All)                   & 0                           & 91.44                                  & 84.59                                  & 4.30                         & 0.99                        \\ \hline
UA-MT~\cite{yu2019uncertainty}                        & MICCAI‘19                 &                            &                             & 46.04                                  & 35.97                                  & 20.08                         & 7.75                         \\
SASSNet~\cite{li2020shape}                      & MICCAI'20                 &                            &                             & 57.77                                   & 46.14                                  & 20.05                         & 6.06                         \\
SS-Net~\cite{wu2022exploring}                       & MICCAI’22                 &                            &                             & 65.83                                  & 55.38                                  & 6.67                         & 2.28                         \\
BCP*~\cite{bai2023bidirectional}                          & CVPR'23                   &                            &                             & 87.61                                  & 78.72                                  & 6.55                         & 1.85                        \\
M-CnT~\cite{huang2024combinatorial}                        & AAAI'24                   &                            &                             & 75.30                                  & -                                      & 10.7                         & -                            \\
LRE~\cite{huang2024exploring}                        & TMI'24                   &                            &                             & 80.49                                  & 69.61                                      & 2.44                         & \textbf{0.59}                            \\
\cellcolor[HTML]{C0C0C0}MA-Sobel & \cellcolor[HTML]{C0C0C0}- &  &  & \cellcolor[HTML]{C0C0C0}88.68 & \cellcolor[HTML]{C0C0C0}80.22 & \cellcolor[HTML]{C0C0C0}2.34 & \cellcolor[HTML]{C0C0C0}0.76 \\
\cellcolor[HTML]{C0C0C0}MA-Canny & \cellcolor[HTML]{C0C0C0}- & \multirow{-7}{*}{3 (5\%)} & \multirow{-7}{*}{67 (95\%)} & \cellcolor[HTML]{C0C0C0}\textbf{88.76} & \cellcolor[HTML]{C0C0C0}\textbf{80.28} & \cellcolor[HTML]{C0C0C0}\textbf{2.28} & \cellcolor[HTML]{C0C0C0}0.63 \\ \hline
UA-MT~\cite{yu2019uncertainty}                        & MICCAI‘19                 &                            &                             & 81.65                                  & 70.64                                  & 6.88                         & 2.02                         \\
SASSNet~\cite{li2020shape}                      & MICCAI'20                 &                            &                             & 84.5                                   & 74.34                                  & 5.42                         & 1.86                         \\
SS-Net~\cite{wu2022exploring}                       & MICCAI’22                 &                            &                             & 86.78                                  & 77.67                                  & 6.07                         & 1.40                         \\
BCP*~\cite{bai2023bidirectional}                          & CVPR'23                   &                            &                             & 89.07                                  & 80.94                                  & 4.23                         & 1.11                         \\
M-CnT~\cite{huang2024combinatorial}                        & AAAI'24                   &                            &                             & 88.40                                  & -                                      & 4.40                         & -                            \\
ABD~\cite{chi2024adaptive}                          & CVPR'24                   &                            &                             & 89.81                                  & 81.95                                  & 1.46                & 0.49                \\
LRE~\cite{huang2024exploring}                        & TMI'24                   &                            &                             & 89.49                                  & 81.53                                      & \textbf{1.42}                         & \textbf{0.46}                \\

\cellcolor[HTML]{C0C0C0}MA-Sobel & \cellcolor[HTML]{C0C0C0}- & & & \cellcolor[HTML]{C0C0C0}89.82 & \cellcolor[HTML]{C0C0C0}82.10 & \cellcolor[HTML]{C0C0C0}3.24 & \cellcolor[HTML]{C0C0C0}0.87 \\
\cellcolor[HTML]{C0C0C0}MA-Canny & \cellcolor[HTML]{C0C0C0}- & \multirow{-7}{*}{7 (10\%)} & \multirow{-7}{*}{63 (90\%)} & \cellcolor[HTML]{C0C0C0}\textbf{89.96} & \cellcolor[HTML]{C0C0C0}\textbf{82.25} & \cellcolor[HTML]{C0C0C0}1.60 & \cellcolor[HTML]{C0C0C0}0.54 \\ \hline
\end{tabular}
}
\end{table*}

\subsection{Encoder}
Currently, the most commonly used architecture for medical image segmentation~\cite{bai2023bidirectional, wang2023mcf} is U-Net~\cite{ronneberger2015u}, and for 3D datasets, the corresponding model is V-Net~\cite{milletari2016v}. Both of these architectures follow an encoder-decoder structure. The encoder processes input data and projects it into feature spaces of varying sizes. Typically, the encoder is composed of a combination of convolutional layers~\cite{o2015introduction} (or 3D convolutional layers~\cite{ravi2020accelerating}), normalization layers, and activation functions. As the data passes through deeper layers, the spatial resolution of the features is progressively reduced while the feature channels are enriched. These feature maps are then passed into the decoder for further processing.

In the encoder stage, we align with the baseline method~\cite{bai2023bidirectional, wang2023mcf} by simultaneously processing both labeled and unlabeled data. We define the 2D image encoder as $\mathbb{E}_{2D}$ and the 3D image encoder as $\mathbb{E}_{3D}$, which handle 2D images $\textbf{I}$ and 3D volumes $\textbf{V}$, respectively. Specifically, we concatenate the labeled and unlabeled inputs batch-wise before feeding them into the encoder to extract features. For 2D images, the extracted features are denoted as $F_I = \mathbb{E}_{2D}(I)$, while for 3D volumes, the features are represented as $F_V = \mathbb{E}_{3D}(V)$. This unified approach ensures that both types of inputs contribute to feature extraction in a cohesive manner.

\subsection{Dual-Branch Decoder}
Unlike existing methods~\cite{tarvainen2017mean, wu2022exploring, wu2022exploring} that employ a single-branch decoder, we introduce a manifold branch in addition to the base branch. These are referred to as the base branch ($\mathbb{D}_{ba}$) and the manifold branch ($\mathbb{D}_{mf}$), respectively. Both $\mathbb{D}_{ba}$ and $\mathbb{D}_{mf}$ receive input features $\textbf{F}$ (either $F_I$ or $F_V$) extracted by the encoder.

Specifically, for the 2D dataset, the base branch $\mathbb{D}_{ba}$ adopts the design of the decoder architecture from the baseline method~\cite{bai2023bidirectional, wang2023mcf, gao2023correlation}, where the final segmentation predictions are generated by the segmentation head $\mathbb{H}_{seg}$, with the number of output channels set to $K$, corresponding to the number of classes in the dataset. In contrast, the manifold branch $\mathbb{D}_{mf}$ follows a U-Net-like architecture, consisting of convolutional layers, normalization layers, activation functions, and upsampling layers. The final manifold predictions are produced by the manifold head $\mathbb{H}_{mf}$, with the number of output channels set to 2.
The outputs from both branches are batch-wise split into labeled and unlabeled predictions, after which the corresponding loss is calculated based on the respective supervision signals.

During training, the encoder and the dual-branch decoders ($\mathbb{D}_{ba}$ and $\mathbb{D}_{mf}$) are trained simultaneously, allowing both branches to optimize the encoder. However, during inference, only the optimized encoder and $\mathbb{D}_{ba}$ are required for the segmentation task. This ensures that, compared to the baseline method, there is no increase in parameters or computational cost during the inference phase.



\subsection{Manifold Generator}
We employ a Manifold Generator, denoted as $\mathbb{M}$, to generate supervision signals for the manifold branch. The input to $\mathbb{M}$ is segmentation supervision signals $\textbf{Y}$, and its output is the manifold supervision signals $\textbf{M}$. Regardless of whether the data is labeled or unlabeled, we apply the same $\mathbb{M}$ for processing. The difference lies in the source of the segmentation supervision: for labeled data, the supervision signals are accurate ground truth, while for unlabeled data, the signals are derived from pseudo-labels generated by the pseudo-label generation network.

The manifold information essentially represents the variations in local gradients of the data, and a variety of established manifold filters~\cite{4767851, sobel19683x3, kovalevsky2022edge, 6100495} can be utilized in designing $\mathbb{M}$. In this paper, we introduce two specific functions for $\mathbb{M}$: the Sobel Operator~\cite{sobel19683x3}, which is applicable to both 2D and 3D data, and the Canny Operator~\cite{4767851}, which is specialized for 2D images. Based on the choice of $\mathbb{M}$, we present two variants of our method: MA-Sobel and MA-Canny.


\subsubsection{Sobel Operator}
The Sobel operator can be applied to both 2D and 3D data. For clarity, we first convert the one-hot 2D segmentation map $\textbf{Y} \in \{0,1,\dots,K-1\}^{W\times H}$ into a discrete categorical representation $\overline{\textbf{Y}} \in \mathbb{R}^{W \times H \times 1}$. We then conceptualize a $3 \times 3$ patch of $\overline{\textbf{Y}}$ as a 3D tensor $\textbf{P} \in \mathbb{R}^{3 \times 3 \times 1}$. 
The Sobel operator uses two $3 \times 3$ kernels to convolve with the input patch $\textbf{P}$ to approximate the horizontal and vertical derivatives~\cite{sobel19683x3}, generating gradient components $\boldsymbol{G_x}$ and $\boldsymbol{G_y}$ according to the following equations:
\begin{equation}
    \begin{aligned}
\bold{G_{x}} = \begin{bmatrix}
-1 & 0 & 1 \\
-2 & 0 & 2 \\
-1 & 0 & 1 
\end{bmatrix}*\textbf{P},~~~
\bold{G_{y}} = \begin{bmatrix}
1 & 2 & 1 \\
0 & 0 & 0 \\
-1&-2 & -1 
\end{bmatrix}*\textbf{P},
    \end{aligned}
    \label{eq:Gx}
\end{equation}
where $\boldsymbol{G_x}, \boldsymbol{G_y} \in \mathbb{R}^1$ represent the approximated gradients. The manifold value at the center of this patch is then derived by approximating the gradient magnitude:
\begin{equation}
    \begin{aligned}
        \textbf{G}=\sqrt{\bold{G_{x}}^2+\bold{G_{y}}^2}.       
    \end{aligned}
     \label{eq:G}
\end{equation}

Therefore, for the entire segmentation map \textbf{Y}, the manifold \textbf{M} generated by the manifold generator is expressed as:
\begin{equation}
    \begin{aligned}
    \textbf{M} = \mathbb{M} (\textbf{Y}) = Sobel(\overline{\textbf{Y}}).
\end{aligned}
     \label{eq:M}
\end{equation}

\begin{table*}[]
\centering
\caption{\label{tab:sota}
	Performance comparison with state-of-the-art methods on LA dataset. 
 }
 \scalebox{0.8}{
\begin{tabular}{c|c|cc|cccc}
\hline
                             &                           & \multicolumn{2}{c|}{Scans Used}                           & \multicolumn{4}{c}{Metrics}                                                                                                                                     \\ \cline{3-8} 
\multirow{-2}{*}{Method}     & \multirow{-2}{*}{Venue}   & Labeled                    & Unlabeled                   & Dice(\%)↑                              & Jaccard(\%)↑                           & 95HD↓                                 & ASD↓                                  \\ \hline
V-Net~\cite{milletari2016v}                        & -                         & 4 (5\%)                    & 0                           & 52.55                                  & 39.6                                   & 47.05                                 & 9.87                                  \\
V-Net~\cite{milletari2016v}                        & -                         & 8 (10\%)                   & 0                           & 82.74                                  & 71.72                                  & 13.35                                 & 3.26                                  \\
V-Net~\cite{milletari2016v}                        & -                         & 80 (All)                   & 0                           & 91.47                                  & 84.36                                  & 5.48                                  & 1.51                                  \\ \hline
UA-MT~\cite{yu2019uncertainty}                        & MICCAI'19                 &                            &                             & 82.26                                  & 70.98                                  & 13.71                                 & 3.82                                  \\
SASSNet~\cite{li2020shape}                      & MICCAI'20                 &                            &                             & 81.6                                   & 69.63                                  & 16.16                                 & 3.58                                  \\
DTC~\cite{luo2021semi}                          & AAAI'21                   &                            &                             & 81.25                                  & 69.33                                  & 14.9                                  & 3.99                                  \\
SS-Net~\cite{wu2022exploring}                       & MICCAI’22                 &                            &                             & 86.33                                  & 76.15                                  & 9.97                                  & 2.31                         \\
CAML~\cite{gao2023correlation}                         & MICCAI’23                 &                            &                             & 87.34                                  & 77.65                                  & 9.76                                  & 2.49                                  \\
BCP*~\cite{bai2023bidirectional}                          & CVPR'23                   &                            &                             & 87.35                                  & 77.76                                  & 8.69                                  & 2.45                                  \\
LRE~\cite{huang2024exploring}                        & TMI'24                   &                            &                             & 87.35                                  & 77.66                                      & 8.96                         & \textbf{2.02}                             \\
\cellcolor[HTML]{C0C0C0}MA-Sobel & \cellcolor[HTML]{C0C0C0}- & \multirow{-7}{*}{4 (5\%)}  & \multirow{-7}{*}{76 (95\%)} & \cellcolor[HTML]{C0C0C0}\textbf{88.03} & \cellcolor[HTML]{C0C0C0}\textbf{78.80} & \cellcolor[HTML]{C0C0C0}\textbf{8.51} & \cellcolor[HTML]{C0C0C0}2.32          \\ \hline
UA-MT~\cite{yu2019uncertainty}                        & MICCAI'19                 &                            &                             & 87.79                                  & 78.39                                  & 8.68                                  & 2.12                                  \\
SASSNet~\cite{li2020shape}                      & MICCAI'20                 &                            &                             & 87.54                                  & 78.05                                  & 9.84                                  & 2.59                                  \\
DTC~\cite{luo2021semi}                          & AAAI'21                   &                            &                             & 87.51                                  & 78.17                                  & 8.23                                  & 2.36                                  \\
MTCL~\cite{9779756}                          & TMI'24                   &                            &                             & 88.34                                  & 79.27                                  & 10.85                                 & 2.36                                     \\
SS-Net~\cite{wu2022exploring}                       & MICCAI’22                 &                            &                             & 88.55                                  & 79.62                                  & 7.49                                  & 1.90                                  \\
CAML~\cite{gao2023correlation}                         & MICCAI’23                 &                            &                             & 89.62                                  & 81.28                                  & 8.76                                  & 2.02                                  \\
BCP*~\cite{bai2023bidirectional}                          & CVPR'23                   &                            &                             & 89.94                                  & 81.81                                  & 6.93                                  & 1.76                                  \\
LRE~\cite{huang2024exploring}                        & TMI'24                   &                            &                             & 89.25                                  & 80.68                                      & 6.96                         & 1.86                             \\
\cellcolor[HTML]{C0C0C0}MA-Sobel & \cellcolor[HTML]{C0C0C0}- & \multirow{-7}{*}{8 (10\%)} & \multirow{-7}{*}{72 (90\%)} & \cellcolor[HTML]{C0C0C0}\textbf{90.28} & \cellcolor[HTML]{C0C0C0}\textbf{82.37} & \cellcolor[HTML]{C0C0C0}\textbf{6.49} & \cellcolor[HTML]{C0C0C0}\textbf{1.66} \\ \hline
\end{tabular}
}
\end{table*}

This 2D Sobel operation can be extended to 3D with minimal modifications. For 3D data, we similarly convert the segmentation map $\textbf{Y}$ into its discrete representation $\overline{\textbf{Y}}$. A patch $\textbf{P} \in \mathbb{R}^{3 \times 3 \times 3}$ is then used, and the 3D Sobel filter consists of three $3 \times 3 \times 3$ kernels, as introduced in~\cite{sobel3d}. The following matrix defines the kernel for the x-axis:
\begin{equation}
    \begin{aligned}
\begin{bmatrix}
-1 & -3 & -1 \\
-3 & -6 & -3 \\
-1 & -3 & -1 
\end{bmatrix}~~~~~~
\begin{bmatrix}
0 & 0 & 0 \\
0 & 0 & 0 \\
0 & 0 & 0 
\end{bmatrix}~~~~~~
\begin{bmatrix}
1 & 3 & 1 \\
3 & 6 & 3 \\
1 & 3 & 1 
\end{bmatrix}
    \end{aligned}
\end{equation}
By rotating this x-axis kernel, we can obtain kernels for the y-axis and z-axis. The process for calculating gradients in 3D data follows the same principles as the 2D Sobel operation.

\subsubsection{Canny Operator}

The Canny Operator algorithm~\cite{4767851} is specifically designed for detecting edges in 2D image manifolds, achieving minimal error and noise. Since edge detection is highly sensitive to noise, the first step of the Canny Operator is to smooth the image using a Gaussian filter. The Gaussian filter~\cite{guassianfilter} is based on a 2D Gaussian function with both x and y dimensions, where the standard deviations in both dimensions are typically set to the same value. The Gaussian function is expressed as:

\begin{equation}
    \begin{aligned}
    Gauss(x, y) = \frac{1}{2\pi\sigma^2} \exp\left(-\frac{x^2 + y^2}{2\sigma^2}\right).
    \end{aligned}
\end{equation}

Here, $\sigma$ is the standard deviation, which controls the degree of smoothing. For instance, a $3 \times 3$ Gaussian kernel with a standard deviation of 1 is commonly used.

Once the Gaussian kernel is defined, it is convolved with the segmentation map \textbf{Y} in a discrete manner to smooth it, we define the smooth map as $\widetilde{\textbf{Y}}$. After smoothing, the next step involves using the Sobel operator to compute the approximate gradient of each pixel. As discussed in the previous section, the gradient in various directions is calculated using Equation~\ref{eq:Gx}, and the manifold $\textbf{M}$ of the smoothed image is obtained using Equation~\ref{eq:G} and~\ref{eq:M}:
\begin{equation}
    \begin{aligned}
    \textbf{M} = \mathbb{M} (\textbf{Y}) = Sobel(\widetilde{\textbf{Y}})
\end{aligned}
     \label{eq:M}
\end{equation}

Finally, a double-thresholding technique is applied to classify pixels into strong edges, weak edges, and non-edges. Let $T_{high}$ and $T_{low}$ represent the high and low thresholds, respectively. Pixels with gradient values greater than $T_{high}$ are categorized as strong edges. Those with gradient magnitudes between $T_{high}$ and $T_{low}$ are considered weak edges, while pixels with gradient values below $T_{low}$ are suppressed as non-edges. These parameters, such as the threshold values, are hyperparameters and can be fine-tuned, with specific settings provided in Section~\ref{sec:ex}.

\vspace{-0.3cm}
\subsection{Pseudo-label Generation Network}
Generating appropriate supervision for unlabeled data is one of the most important components of semi-supervised learning~\cite{basak2023pseudo}. Low-quality pseudo-labels significantly affect the overall model performance. The method proposed in this paper is designed to integrate seamlessly with baseline approaches that utilize pseudo-labels, leveraging the pseudo-label generator from the baseline directly.

Some commonly used techniques for generating pseudo-labels are Mean-Teacher framework~\cite{yu2019uncertainty} and cross-pseudo supervision~\cite{chen2021semi}. The mean-teacher framework typically consists of two identical networks: a teacher and a student network. The student network updates the parameter of the teacher network through the exponential moving average. The output of the teacher network is used as the pseudo-label of the student network. Cross-pseudo supervision also consists of two networks. Unlike the previous method, cross-pseudo supervision uses the pixel-wise one-hot label map from one network to supervise the pixel-wise confidence map of the other network. To demonstrate the versatility of our method, we adopt both of these pseudo-label generation techniques as baseline methods~\cite{bai2023bidirectional, wang2023mcf, gao2023correlation} in the experimental section.

\subsection{Loss Function}

The overall loss function of the network, \( L_{total} \), consists of two main components: \( L_{base} \), which originates from segmentation supervision, and \( L_{mf} \), which arises from manifold supervision. Specifically, \( L_{base} \) employs a loss function consistent with the baseline method~\cite{bai2023bidirectional, wang2023mcf, gao2023correlation}, with one of the most commonly used loss functions being the Dice Loss~\cite{sudre2017generalised}. It is computed as follows:
\begin{equation}
    \begin{aligned}
        L_{base} = DICE(\hat{Y}^{l}, Y^{l}) + DICE(\hat{Y}^{p}, Y^{p}_{pseudo})
    \end{aligned}
\end{equation}

Here, $DICE(*, *)$ represents the Dice coefficient-based loss function, where \( \hat{Y}^{l} \) and \( Y^{l} \) correspond to the predicted and ground truth segmentation maps for labeled data, respectively, while \( \hat{Y}^{p} \) and \( Y^{p}_pseudo \) refer to the predicted and pseudo-label maps for unlabeled data. 

On the other hand, the task of learning manifold information can be viewed as a binary classification problem, where each pixel/voxel is classified as either 0 or 1. As such, the cross-entropy loss (CE) is adopted in \( L_{mf} \) to supervise this process. The calculation of \( L_{mf} \) is as follows:

\[
L_{mf} = CE(\hat{M}^{l}, M^{l}) + CE(\hat{M}^{p}, M^{p}_{pseudo})
\]

Here, \( \hat{M}^{l} \) and \( M^{l} \) represent the predicted and ground truth manifold maps for labeled data, respectively, while \( \hat{M}^{u} \) and \( M^{u}_{pseudo} \) correspond to the predicted and pseudo-labeled manifold maps for unlabeled data. The cross-entropy loss, a standard metric for classification tasks, measures the dissimilarity between predicted probabilities and the actual labels, ensuring the manifold branch accurately learns to distinguish different points categories.

By combining \( L_{base} \) and \( L_{mf} \), the total loss function, \( L_{total} \), effectively balances the supervision from both segmentation and manifold learning tasks, enhancing the network's ability to generalize from both labeled and unlabeled data. The final formulation is:

\[
L_{total} = (1-\alpha) \cdot L_{base} + \alpha \cdot L_{mf}
\]

where \( \cdot \) is weighting factor that balance the contribution of the base and manifold branches to the network's overall optimization objective. This design allows for an adaptive balance between segmentation accuracy and feature learning across both labeled and unlabeled data.
\section{Experiments}
\label{sec:ex}

\noindent
\textbf{Datasets.}

(1) Automatic Cardiac Diagnosis Challenge (ACDC)~\cite{bernard2018deep}. ACDC dataset contains 100 2D cardiac magnetic resonance images (CMRI) with four classes including background, right ventricle, left ventricle, and myocardium.

(2) Left Atrium Dataset (LA)~\cite{xiong2021global}. Atrial Segmentation Challenge dataset includes 100 3D late gadolinium-enhanced magnetic resonance image (LGE-MRI) with an isotropic resolution of 0.625 $\times$ 0.625 $\times$ 0.625 $mm^3$ and corresponding ground truth labels.

(3) Pancreas-NIH~\cite{roth2015deeporgan}. Pancreas-NIH dataset contains 82 contrasts-enhanced abdominal 3D CT volumes with manual annotations.

\noindent
\textbf{Evaluation Metrics.}
For quantitative evaluation, we employed 4 widely used metrics, including \textit{Dice Score} (\%), \textit{Jacard Score} (\%),  \textit{95\% Hausdorff Distance (95HD)}, and \textit{Average Surface Distance (ASD)}. Dice and Jacard Score mainly measure the overlap between two object regions. 95HD measures the closest point distance between two object regions and ASD computes the average distance between their boundaries.

\noindent
\textbf{Implementation Details.}
We conduct all experiments on an NVIDIA A100 GPU. We adopted several approaches~\cite{bai2023bidirectional, wang2023mcf, gao2023correlation} as our baseline networks to showcase our proposed method's effectiveness and generalization capability. We strictly follow the original setting of respective baseline networks to eliminate factors that may influence the model performance. In all the settings, we only added the manifold branch without altering any other settings. This ensures that the performance improvements obtained are attributed to the application of manifold information, rather than other factors. For Canny Operator implementation, $\sigma = 1$ for Gaussian Filter, and $T_{low} = 0.1$, $T_{high} = 0.2$ for thresholding. Detailed implementations of our method can be found in the Supplementary Materials. 

\begin{table}[h!]
\caption{\label{tab:Pancreas}
    	Performance comparison with state-of-the-art methods on Pancreas-NIH dataset. Following CoraNet~\cite{shi2021inconsistency} Labeled: 12 (20\%) Unlabeled: 50 (80\%)
 }
\scalebox{0.75}{
\centering
\begin{tabular}{c|c|cccc}
\hline
                         &                         & \multicolumn{4}{c}{Metrics}                                     \\ \cline{3-6} 
\multirow{-2}{*}{Method} & \multirow{-2}{*}{Venue} & Dice(\%)↑      & Jaccard(\%)↑   & 95HD↓         & ASD↓          \\ \hline
V-Net~\cite{milletari2016v}                    & -                       & 69.96          & 55.55          & 14.27         & 1.64          \\
UA-MT~\cite{yu2019uncertainty}                    & MICCAI'19               & 77.26          & 63.82          & 11.90         & 3.06          \\
SASSNet~\cite{li2020shape}                  & MICCAI'20               & 77.66          & 64.08          & 10.93        & 3.05          \\
CoraNet~\cite{shi2021inconsistency}                  & TMI'21               & 79.67          & 66.69          & 7.59        & 1.89          \\
DTC~\cite{luo2021semi}                      & AAAI'21                 & 78.27          & 64.75          & 8.36 & 2.25 \\
BCP*~\cite{bai2023bidirectional}                      & CVPR'23                 & 82.43          & 70.53          & 15.89         & 4.53          \\
LRE~\cite{huang2024exploring}          & TMI'24              & 81.17                                  & 68.68                                      & \textbf{6.17}                         & \textbf{1.46}                             \\
\rowcolor[HTML]{C0C0C0} 
MA-Sobel                     & -                       & \textbf{82.75} & \textbf{70.90} & 11.15         & 3.64          \\ \hline
\end{tabular}
}
\end{table}

\begin{table*}[]
\caption{\label{tab:baseline}
	Performance comparison when applying our methods on different baseline networks on LA dataset. Improvements compared with the baseline methods are \color{blue}{\textbf{highlighted}}.
 }
\centering
\scalebox{0.75}{
\begin{tabular}{c|c|cc|llll}
\hline
                                   &                              & \multicolumn{2}{c|}{Scans Used}                          & \multicolumn{4}{c}{Metrics}                                                                                                                                            \\ \cline{3-8} 
\multirow{-2}{*}{Architecture}     & \multirow{-2}{*}{Method}     & Labeled                    & Unlabeled                   & Dice(\%)↑                                                 & Jaccard(\%)↑                       & 95HD↓                             & ASD↓                              \\ \hline
                                   & Baseline*                     &                            &                             & 89.94                                                     & 81.81                              & 6.93                              & 1.76                              \\
\multirow{-2}{*}{BCP~\cite{bai2023bidirectional} (CVPR'23)}    & \cellcolor[HTML]{C0C0C0}Ours &                            &                             & \cellcolor[HTML]{C0C0C0}{\color[HTML]{333333} 90.28\color{blue}\textbf{↑0.34}} & \cellcolor[HTML]{C0C0C0}82.37\color{blue}\textbf{↑0.56} & \cellcolor[HTML]{C0C0C0}6.49\color{blue}\textbf{↓0.44} & \cellcolor[HTML]{C0C0C0}1.66\color{blue}\textbf{↓0.10} \\
                                   & Baseline*                     &                            &                             & 89.79                                                     & 81.58                              & 6.87                              & 2.08                              \\
\multirow{-2}{*}{CAML~\cite{gao2023correlation} (MICCAI’23)} & \cellcolor[HTML]{C0C0C0}Ours & \multirow{-4}{*}{8 (10\%)} & \multirow{-4}{*}{72 (20\%)} & \cellcolor[HTML]{C0C0C0}90.11\color{blue}\textbf{↑0.32}                        & \cellcolor[HTML]{C0C0C0}82.07\color{blue}\textbf{↑0.49} & \cellcolor[HTML]{C0C0C0}6.60\color{blue}\textbf{↓0.27} & \cellcolor[HTML]{C0C0C0}1.81\color{blue}\textbf{↓0.27} \\ \hline
                                   & Baseline*                     &                            &                             & 89.79                                                     & 81.79                              & 6.01                              & 1.78                              \\
\multirow{-2}{*}{MCF~\cite{wang2023mcf} (CVPR'23)}    & \cellcolor[HTML]{C0C0C0}Ours & \multirow{-2}{*}{16(20\%)} & \multirow{-2}{*}{84 (80\%)} & \cellcolor[HTML]{C0C0C0}90.64\color{blue}\textbf{↑0.85}                        & \cellcolor[HTML]{C0C0C0}83.02\color{blue}\textbf{↑1.23} & \cellcolor[HTML]{C0C0C0}5.62\color{blue}\textbf{↓0.39} & \cellcolor[HTML]{C0C0C0}1.66\color{blue}\textbf{↓0.12} \\ \hline
\end{tabular}
}
\end{table*}

\begin{table}[]
\caption{\label{tab:alpha}
	Comparison of using different ratios between manifold- and segmentation-loss using LA dataset with 10\% labeled data.
 }
\centering
\scalebox{0.85}{
\begin{tabular}{c|cccc}
\hline
\multirow{2}{*}{$\alpha$} & \multicolumn{4}{c}{Metrics}             \\ \cline{2-5} 
                                & Dice(\%)↑ & Jaccard(\%)↑ & 95HD↓ & ASD↓ \\ \hline
0.1                             & 89.84     & 81.65        & 7.39  & 1.69 \\
0.05                            & \textbf{90.28}     & \textbf{82.37}        & \textbf{6.49}  & \textbf{1.66} \\
0.01                            & 90.08     & 82.03        & 6.71  & 1.67 \\
0.001                           & 89.31     & 80.81        & 7.5   & 1.87 \\
 \hline
\end{tabular}
}
\end{table}

\begin{table}[]
\caption{\label{tab:Pretrain}
	Comparison of applying MANet on either pretrain or self-train using LA dateset with 10\% labeled data.
 }
\scalebox{0.85}{
\centering
\begin{tabular}{cc|cccc}
\hline
\multicolumn{2}{c|}{Training Stage} & \multicolumn{4}{c}{Metrics}             \\ \hline
Pre-train   & Fine-tune  & Dice(\%)↑ & Jaccard(\%)↑ & 95HD↓ & ASD↓ \\ \hline
         &             & 89.94                                                     & 81.81                              & 6.93                              & 1.76  \\
\checkmark         &             & 90.04     & 81.98        & 6.91  & 1.66 \\
           & \checkmark           & 89.81     & 81.61        & 6.86  & 1.73 \\
\checkmark          & \checkmark           & \textbf{90.28}     & \textbf{82.37}        & \textbf{6.49}  & \textbf{1.66} \\ \hline
\end{tabular}
}
\end{table}

\subsection{Experiments on Different Datasets}

\noindent
\textbf{ACDC dataset.} 
We follow existing work~\cite{bai2023bidirectional} and conduct experiments on the dataset using two popular settings: 5\% and 10\% labeled data. The results of our MANet (using the BCP method as the baseline) compared with several state-of-the-art methods~\cite{yu2019uncertainty,li2020shape,luo2021semi,bai2023bidirectional, chi2024adaptive, huang2024combinatorial}, are shown in Table~\ref{tab:ACDC}.
As illustrated in the table, our method achieves optimal performance in four metrics with 88.76\%, 80.28\%, 2.28, and 0.63 when using 5\% labeled data. When 10\% labeled data is used, MA-Canny also achieves 89.96\% in Dice and 82.25\% in Jaccard score. 

We applied both MA-Sobel and MA-Canny to the ACDC dataset, a 2D dataset. While both variants demonstrated effective performance improvements compared to the baseline method (BCP), a noteworthy observation is that MA-Canny consistently outperformed MA-Sobel across both 5\% and 10\% labeled data settings. This improvement was especially pronounced in boundary-related metrics, such as the 95HD and ASD. Specifically, in the 10\% labeled data setting, MA-Canny reduced 95HD and ASD by 1.64 and 0.33, respectively, compared to MA-Sobel.
This suggests that while MA-Sobel is more versatile and applicable to both 2D and 3D datasets, MA-Canny demonstrates superior capability for 2D data, likely due to the precision of the Canny operator in detecting finer boundaries compared to the Sobel operator. These results indicate that MA-Canny is better suited for 2D datasets like ACDC, where more accurate boundary detection is critical. Furthermore, it implies that the performance of MANet could be further enhanced with more advanced manifold generation techniques, emphasizing the potential for continual improvement.

\noindent
\textbf{LA dataset.}
We adopted two popular settings for this dataset, utilizing 5\% and 10\% of the labeled data. The results of our MANet (using the BCP method as the baseline) in comparison with several recently developed methods~\cite{yu2019uncertainty,li2020shape,luo2021semi,gao2023correlation,bai2023bidirectional} are presented in Table~\ref{tab:sota}. As illustrated in the table, our method achieved favorable model performance, with Dice and Jaccard scores of 88.03\% and 78.80\%, respectively, using 5\% labeled data. Notably, when using 10\% labeled data, our method attained a Dice score of 90.28\%, which is very close to the fully-supervised benchmark method V-Net performance, i.e., 91.47\% (which uses 100\% labeled data). This demonstrates the potential of MANet as a semi-supervised model.


\noindent
\textbf{Pancreas-NIH dataset.}
To ensure a fair comparison with other methods~\cite{yu2019uncertainty,li2020shape,bai2023bidirectional, luo2021semi, shi2021inconsistency}, we utilized 20\% of the labeled data from this dataset, consistent with the other approaches. The results, presented in Table~\ref{tab:Pancreas}, demonstrate that our method (using the BCP method as the baseline) outperforms all other competitors in terms of Dice and Jaccard scores. As shown in Table~\ref{tab:Pancreas}, our approach achieves superior performance, highlighting its effectiveness.

\subsection{Experiments on Different Architectures}

We propose MANet, a versatile architecture for medical image semantic segmentation. The manifold branch of MANet can be easily integrated into existing methods as an additional decoder branch within the U-Net structure, enhancing the performance of baseline methods. To validate its generalizability, we evaluated our approach on the LA dataset using different state-of-the-art semi-supervised medical image semantic segmentation architectures (e.g., BCP~\cite{bai2023bidirectional}, MCF~\cite{wang2023mcf} and CAML~\cite{gao2023correlation}). The results, reported in Table~\ref{tab:baseline}, demonstrate that incorporating the MANet structure consistently leads to significant performance improvements across all settings, thereby proving the generalizability of our method.

\subsection{Visualization}

\begin{figure*}[t!]
	\centering
	\centering
	\includegraphics[width=0.95\textwidth]{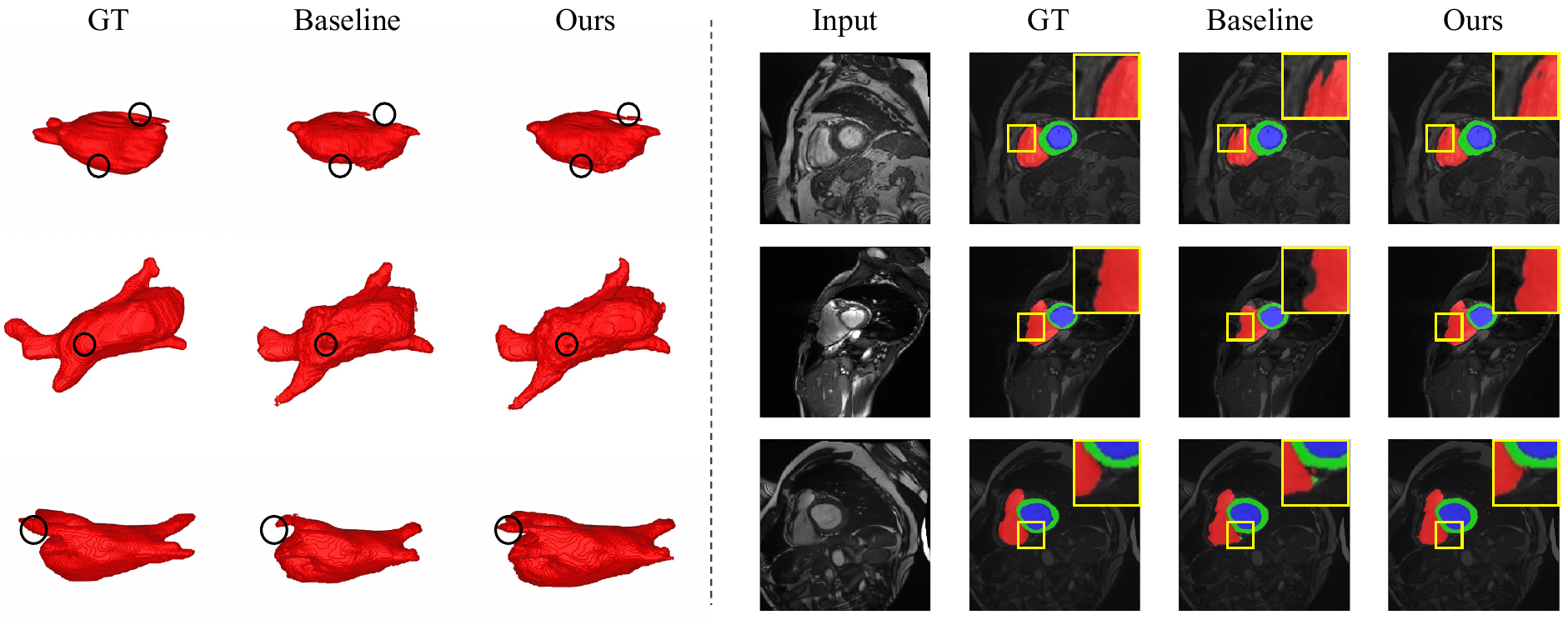}
        \vspace{-0.3cm}	
	\caption{The visualization results from the LA dataset (left) and the ACDC dataset (right), with both datasets using 10\% labeled data. The baseline method employed is  BCP~\cite{bai2023bidirectional}.}
	\label{fig:vis}
\end{figure*}

Figure~\ref{fig:vis} illustrates the qualitative results of our method on the LA (3D) and ACDC (2D) datasets, using BCP as the baseline method.  
For the LA dataset, the boundary regions of the organs are more accurately reproduced with our method. For instance, in the 1st and 3rd rows, the baseline segmentation results show missing protruding parts, whereas our method successfully completes these segments. In the 2nd row, there is a noticeable indentation in the baseline segmentation result's boundary, which is nearly eliminated in our segmentation result. Similar effects can be observed in the ACDC dataset. In the 1st and 2nd rows, the baseline segmentation results exhibit missing boundary pixels. With the assistance of manifold branch, the model captures local details and refines the boundaries. In the last row, our method corrects the misclassified pixels and the protruding corner present in the baseline results. 
These findings are consistent with our observations and motivations presented in this paper, demonstrating that integrating these two types of supervision signals during training benefits the modeling of local features.

\subsection{Ablation Studies}
\noindent
\textbf{Different values of weighting parameter $\alpha$.} 
The loss function in our proposed method relies on a weighting parameter, $\alpha$, which controls the loss between the two branches. To balance the segmentation supervision and manifold supervision signals, we set $\alpha$ as a hyperparameter and conducted experiments with different $\alpha$ values (including 0.1, 0.05, 0.01, and 0.001) to analyze their impact on model performance. The results shown in Table~\ref{tab:alpha} indicate that the value of $\alpha$ indeed affects the model's performance. After carefully weighing the performance of all metrics, we adopt $\alpha$=0.05 as our final choice.
 
\noindent
\textbf{MANet in different training stage.} 
Many existing methods~\cite{bai2023bidirectional, yu2019uncertainty, xie2020self} pre-train networks using labeled data and then fine-tune the network parameters using both labeled and unlabeled data. Pre-trained networks not only provide a good initialization for subsequent training but can also serve as pseudo-label generators~\cite{xie2020self} or teacher networks~\cite{bai2023bidirectional, yu2019uncertainty}. As a versatile structure, MANet can be applied to different training stages. To evaluate the performance of MANet at different training stages, we conducted experiments and presented the results in Table~\ref{tab:Pretrain}. From the table, we can observe the following findings: 1) Even when MANet is only used during the pre-training stage and the baseline method is used during the fine-tuning stage, the model's final performance improves. This demonstrates that MANet can independently provide better initialization parameters during the pre-training stage.
2) When MANet is only used during the fine-tuning stage, the performance decreases. We speculate that this is because MANet is randomly initialized during the self-training stage, making it more challenging to converge compared to other parameters that already have a good initialization.
3) When MANet is used in both the two stages, the performance reaches its best.

\label{sec:con}
\section{Conclusion}

In this paper, we propose a novel Manifold-Aware Local Feature Modeling Network (MANet) to address the challenge of improving boundary accuracy in medical image segmentation. 
MANet outperforms existing methods on ACDC, LA, and Pancreas-NIH datasets, improving Dice and Jaccard scores. 
It offers versatile semi-supervised capabilities without increasing network parameters or inference time. 
Additionally, we introduce two variants of MANet: MA-Sobel and MA-Canny. MA-Sobel, which utilizes the Sobel operator, is applicable to both 2D and 3D data, making it a general solution across various datasets. MA-Canny, on the other hand, employs the Canny operator specifically for 2D images. 
Our experiments show that MA-Canny delivers superior performance on 2D datasets compared to MA-Sobel, demonstrating its effectiveness in refining boundary detection when applied to 2D medical images.
Furthermore, MANet can be utilized independently during the pre-training phase, providing superior initialization parameters for subsequent training without increasing network parameter count or inference time during the testing phase. However, this approach also has limitations, such as dependence on the quality of pseudo-labels and increased computational complexity during training. Future work could explore integrating this operation into the pseudo-label generation process to produce higher-quality pseudo-labels and potentially achieve better performance.

%
\IEEEpeerreviewmaketitle



\ifCLASSOPTIONcaptionsoff
  \newpage
\fi



%

\bibliographystyle{IEEEtran}
\bibliography{main.bib}


\end{document}